%% file: main.tex
\documentclass[letterpaper, 10 pt, conference]{ieeeconf}
\IEEEoverridecommandlockouts
\input{macros}
\begin{document}

\title{Game-Theoretic Planning for Autonomous Driving \\ among Risk-Aware Human Drivers}

\author{Rohan Chandra$^{*1}$,
        Mingyu Wang$^{*2}$,
        Mac Schwager$^{3}$ and 
        Dinesh Manocha$^{1,4}$ 
\thanks{* denotes equal contribution to the work.}
\thanks{Wang and Schwager are supported in part by ONR grant N00014-18-1-2830. Chandra and Manocha are supported by ARO Grants W911NF1910069, W911NF2110026, U.S. Army Grant No. W911NF2120076  and, Semiconductor Research Corporation (SRC) and Intel. Toyota Research Institute provided funds to support this work.}
\thanks{$^{1}$Department of Computer Science, University of Maryland, USA
        {\tt\small \{rchandr1, dmanocha\}@umd.edu}}%
\thanks{$^{2}$Department of Mechanical Engineering, Stanford University, Stanford,USA
        {\tt\small mingyuw@stanford.edu}}%
\thanks{$^{3}$Department of Aeronautics and Astronautics, Stanford University, Stanford, USA
        {\tt\small schwager@stanford.edu}}%
\thanks{$^{4}$Department of Electrical and Computer Engineering, University of Maryland, USA
        {\tt\small schwager@stanford.edu}}%
}

\maketitle

\input{Sections/0-Abstract}
\input{Sections/1-intro}

\input{Sections/2-related}

\input{Sections/3-background}
\input{Sections/4-approach}

\input{Sections/6-experiments}

\input{Sections/7-conclusion}

\bibliography{refs}
\bibliographystyle{plain}


\end{document}

%% file: macros.tex
\newcommand{\mc}[1]{\mathcal{#1}}

\makeatletter
\newcommand\footnoteref[1]{\protected@xdef\@thefnmark{\ref{#1}}\@footnotemark}
\makeatother

\newcommand{\shorteq}{%
  \settowidth{\@tempdima}{-}
  \resizebox{\@tempdima}{\height}{=}%
}

\usepackage{amssymb,fge}

\usepackage{amsmath, amssymb}
\usepackage{pifont}
\usepackage{subcaption}
\usepackage[utf8]{inputenc}
\usepackage[T1]{fontenc}
\usepackage[english]{babel}
\usepackage[linesnumbered,ruled,vlined]{algorithm2e}
\usepackage[font=small,belowskip=-1.5pt]{caption} 
\usepackage{array}
\usepackage{graphicx}
\usepackage{amsfonts}
\usepackage{bm}
\usepackage{soul}
\usepackage{hhline}
\usepackage{multirow, makecell}
\usepackage{float}
\usepackage{booktabs,wrapfig}

\usepackage{amsthm}
\usepackage{color}
\usepackage{transparent}
\usepackage{url}
\usepackage{footmisc}
\usepackage{setspace}
\makeatletter
\let\NAT@parse\undefined
\makeatother
\usepackage[colorlinks=true, citecolor=magenta]{hyperref}
\usepackage{mathtools}
\theoremstyle{plain}


%% file: Sections/0-Abstract.tex
\begin{abstract}

We present a novel approach for risk-aware planning with human agents in multi-agent traffic scenarios. Our approach takes into account the wide range of human driver behaviors on the road, from aggressive maneuvers like speeding and overtaking, to conservative traits like driving slowly and conforming to the right-most lane. In our approach, we learn a mapping from a data-driven human driver behavior model called the CMetric to a driver's entropic risk preference.  We then use the derived risk preference within a game-theoretic risk-sensitive planner to model risk-aware interactions among human drivers and an autonomous vehicle in various traffic scenarios. We demonstrate our method in a merging scenario, where our results show that the final trajectories obtained from the risk-aware planner generate desirable emergent behaviors. Particularly, our planner recognizes aggressive human drivers and yields to them while maintaining a greater distance from them. In a user study, participants were able to distinguish between aggressive and conservative simulated drivers based on trajectories generated from our risk-sensitive planner. We also observe that aggressive human driving results in more frequent lane-changing in the planner. Finally, we compare the performance of our modified risk-aware planner with existing methods and show that modeling human driver behavior leads to safer navigation.

\end{abstract}

%% file: Sections/1-intro.tex
\section{Introduction}
\label{sec: introduction}

Risk-aware planning involves sequential decision-making in dynamic and uncertain environments, where agents must consider the risks associated with their actions and corresponding costs and rewards~\cite{cai2021risk}. \textit{Risk-seeking} agents are willing to take lower expected reward in exchange for a higher reward variance (more risk), while \textit{risk-averse} agents are willing to take a lower expected reward in exchange for lower reward variance (less risk). Agents that are risk-averse or risk-seeking are collectively referred to as risk-aware. Human drivers are risk-aware by nature~\cite{majumdar2017risk,ratliff2019inverse,kwon2020humans}. For example, aggressive drivers frequently speed, overtake, and perform sharp cut-ins, whereas conservative drivers drive more cautiously. To navigate successfully among human drivers, autonomous vehicles (AVs) must identify the risk preferences of human drivers online, and predict and plan future motion with the risk preferences of all agents in mind, including the AV's own risk preferences. 

The most common risk measures utilized in risk-sensitive planning are entropic risk~\cite{follmer2011entropic} and conditional value at risk (CVaR)~\cite{rockafellar2002conditional}. A popular approach for risk-aware planning in multi-agent traffic scenarios is to model risk-aware agent interactions via dynamic games~\cite{wang2020game} wherein agents act while considering their impact on other agents as well as the intentions of the other agents. In~\cite{wang2020game}, the authors compute the Nash equilibrium solution of the game by iteratively solving a set of LEQ equations~\cite{whittle1990risk,fleming1992risk}. The main benefits of this approach compared to prior risk-aware planning methods include improved time-to-goal and, more importantly, generation of emergent behaviors. For instance, risk-averse agents learn to maintain a greater distance from risk-seeking agents and generally yield more frequently to risk-seeking agents at intersections, at roundabouts, and during merging.

\begin{figure}[t]
    \centering
    \includegraphics[width=\columnwidth]{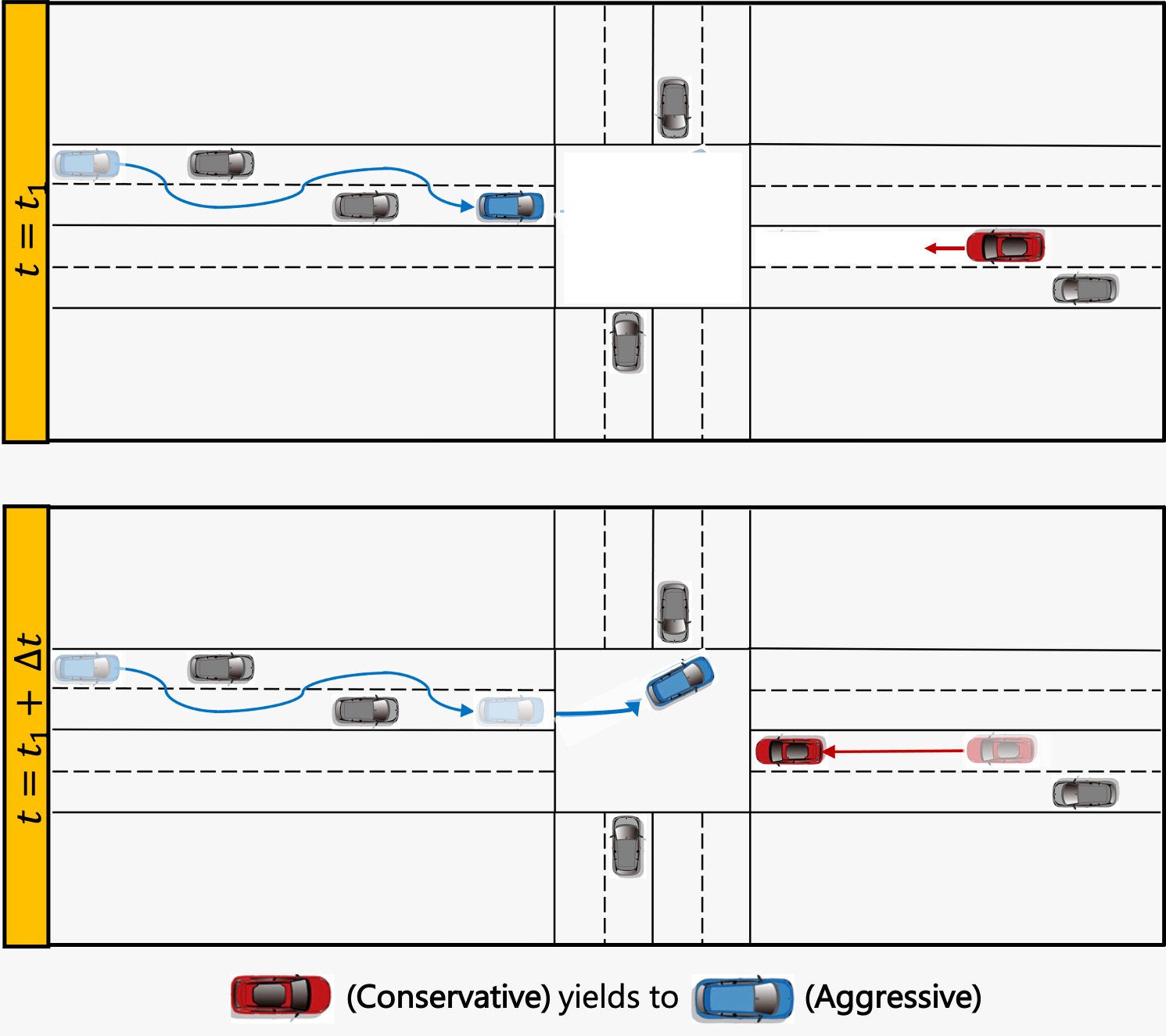}
    \caption{\textbf{Risk-aware planning with human agents:} We present a novel risk-aware planner that takes into account human driver behavior. In the first step (\textit{top}), we characterize the human agent as aggressive (blue agent) with the ego-vehicle as a conservative agent(red). In the second step (\textit{bottom}), we derive the corresponding risk sensitivity of the human agent and generate a game-theoretically optimal and safe risk-aware trajectory for the red agent that suggests the red agent yields to the blue human driver allowing it to cross first). }
    \label{fig: cover}
    \vspace{-10pt}
\end{figure}

Despite its performance and benefits, the main drawback of the approach proposed by~\cite{wang2020game} is that it does not model the risk tolerance of human drivers, as it assumes the AV knows the synthetically chosen risk tolerances for all other driving agents). Extending game-theoretic risk-aware planning to human drivers will allow AVs to act more confidently around human drivers and reduce time-to-goal via more efficient and safer navigation. Estimating the risk tolerances of human drivers, however, requires computationally tractable human driver behavior models that can characterize drivers. 

Some of the state-of-the-art approaches for modeling human driver behavior~\cite{rabinowitz2018machine, jara2019theory, schwarting2019social, gt2} are data-driven and require a large volume of clean training data. These methods classify behaviors as aggressive and conservative~\cite{rabinowitz2018machine, jara2019theory} or selfish and altruistic~\cite{schwarting2019social}. In contrast, deterministic models~\cite{cmetric} do not require data and assign a real-valued score to each agent to quantify its behavior. These approaches can be integrated with risk-aware planning frameworks to incorporate planning for human agents~\cite{suriyarachchi2022gameopt, chandra2022gameplan, bgap}.

\noindent\textbf{Main Contributions:}
We propose a novel approach for risk-aware planning in multi-agent traffic scenarios that takes into account human driver behaviors. We extend an existing risk-aware planner~\cite{wang2020game} by incorporating interactions with human drivers using a data-driven human driver behavior model~\cite{cmetric}. We derive a linear mapping between the driver behavior and risk tolerance (Equation~\ref{eq: mapping}), which serves as the key component of our proposed approach.

To evaluate our approach, we validate the mapping between driver behavior and risk tolerance by measuring the number of lane changes, and test the accuracy of this model via K-Means clustering. Our results show that aggressive human driving results in more frequent lane changing. We confirm that the final trajectories obtained from the risk-aware planner generate emergent behaviors. We measure the yield \% and minimum distance between human drivers at intersections, at roundabouts, and during merging where we observe that conservative drivers generally yield to aggressive drivers while maintaining a greater distance from them. We also conduct a user study in which we show that users are able to distinguish between aggressive and conservative trajectories generated by the planner.

Finally, we compare our modified risk-aware planner with existing planners that do not model human drivers and show that modeling human drivers results in safer navigation. Specifically,~\cite{wang2020game} (and similar planners) assign a fixed neutral risk tolerance to human drivers and the ego-vehicle generates to the human driver accordingly. However, when the human driver is, in fact, either aggressive or conservative, then we show that the error (absolute value of minimum relative distance between the agents) increases by $10\%$. We explain this in detail in Section~\ref{subsubsec: Comparing_with_baseline}.

%% file: Sections/2-related.tex
\section{Related Work}
\label{sec: related_work}

\subsection{Risk-Aware Planning}
\label{subsec: risk_aware_planning}

Risk sensitivity-based planning~\cite{osogami2012robustness,chow2014framework,chow2015risk, samuelson2018safety,chapman2019risk} considers the risk associated with the actions of agents to avoid unsafe situations. The most common risk measures utilized in risk-sensitive planning are entropic risk~\cite{follmer2011entropic} and conditional value at risk (CVaR)~\cite{rockafellar2002conditional}. Entropic risk has been widely used in optimal control due to its simplicity and 
tractability~\cite{whittle1981risk}, while CVaR has recently been incorporated in trajectory optimization due to its interpretability~\cite{chow2015risk}. Risk-aware planning has been used extensively in autonomous underwater vehicles~\cite{risk-water}, ground vehicles~\cite{risk-ground, risk-ground2}, and unmanned aerial vehicles (UAVs)~\cite{risk-uav}. While the CVaR risk model has been used in the latter two cases,~\cite{risk-water} used the entropic measure of risk. In addition to CVaR and the entropic models, several other models are also used in various applications such as the dynamic risk density function for collision avoidance~\cite{risk-CA} and semantic maps for simultaneous localization and mapping (SLAM)~\cite{risk-semantic1, risk-semantic2}

\subsection{Data-Driven Methods for Driver Behavior Prediction}
\label{subsec: data_driven_related}

Data-driven methods broadly follow two approaches. In the first approach, various machine learning algorithms such as clustering, regression, and classification predict or classify driver behavior as either aggressive or conservative. These methods have been studied in traffic psychology and the social sciences~\cite{taubman2004multidimensional, gulian1989dimensions, french1993decision, deffenbacher1994development, ishibashi2007indices,ernestref2,ernestref3,ernestref4,ernestref8,ernestref9,ernestref10,ernestref11,ernestref12,ernestref12,ernestref13,ernestref14,ernestref15,ernestref16}. So far, there has been relatively little work to improve the robustness and ability to generalize to different traffic scenarios, steps that require ideas from computer vision and robotics.

The second approach uses trajectories to learn reward functions for human behavior using inverse reinforcement learning (IRL)~\cite{rabinowitz2018machine, jara2019theory, schwarting2019social, gt2}. IRL-based methods, however, have certain limitations. IRL requires large amounts of training data, and the learned reward functions are unrealistically tailored towards scenarios only observed in the training data~\cite{rabinowitz2018machine, gt2}. For instance,~\cite{rabinowitz2018machine} requires $32$ million data samples for optimum performance. Additionally, IRL-based methods are sensitive to noise in the trajectory data~\cite{schwarting2019social, gt2}. Consequently, current IRL-based methods are restricted to simple and sparse traffic conditions.

%% file: Sections/3-background.tex
\section{Preliminaries and Problem Formulation}
\label{sec: Preliminaries_and_Problem_Formulation}

In this section, we briefly summarize the 
CMetric algorithm~\cite{cmetric} for behavior modeling. We will also define the problem statement of risk-aware planning.

\input{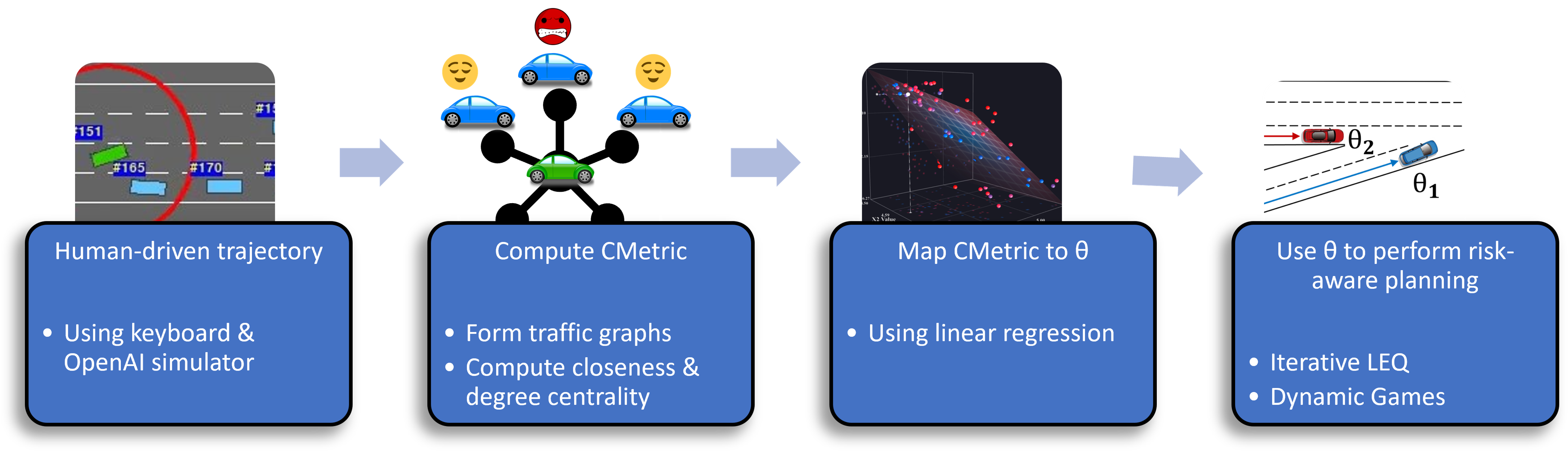}


\subsection{CMetric: Modeling Human Driver Behavior}
\label{subsec: cmetric_background}

We briefly summarize the CMetric algorithm~\cite{cmetric, chandra2021using, chandra2019graphrqi} because it provides an objective measure of aggressiveness based on driving maneuvers such as speeding, overtaking, and so on. To determine if an agent is aggressive or conservative, the algorithm begins by reading the trajectories of the agent and surrounding vehicles via cameras or lidars during a time period $\Delta T$. The trajectory of an agent $i$ is represented by 

\[\Xi^i= \{x^i_t \ | \ t = t_0, t_1, \ldots, t_0 + T \}.\]

\noindent In CMetric, these trajectories are represented via weighted undirected graphs $\mathcal{G} = (\mathcal{V}, \mathcal{E})$ in which the vertices denote the positions for each agent and the edges correspond to the distances between agents. The algorithm proceeds by using these graphs to model the likelihood and intensity of driving behavior indicators like speeding, overtaking, sudden zigzagging, and lane-changes via the closeness and degree centrality functions~\cite{rodrigues2019network} represented by $\Phi: \mathcal{G}\rightarrow \mathbb{R}$. These behavior indicators determine whether an agent is aggressive or conservative~\cite{sagberg2015review}. The behavior profile for agent $i$ is denoted by $\bm{\zeta}_i$ and is computed as,

\begin{equation}
    \bm{\zeta}_i(\Xi_{\Delta t}) = \Phi(\mathcal{G}),
    \label{eq: cmetric}
\end{equation}

\noindent where $\mathcal{G}(\mathcal{V}, \mathcal{E})$ is constructed using $\Xi_{\Delta t}$. 

\subsection{Problem Formulation}
\label{sec: problem_formulation}

We consider $N$ agents consisting of a mixture of human drivers and AVs with the system dynamics defined by Equation~\ref{eq:pre_oc_lineardynamics}, and we define the cost function for each agent by Equation~\ref{eq: risk_cost_function}. A human driver is simulated using a user-controlled keyboard with the following features: acceleration, braking, and lane changing. For simplicity, we test with one human driver, but our approach can work with more than one human driver. We further assume that agents are non-ideal in that agents are not provided the risk tolerance of other agents. The input to our approach consists of the state and control signals of every agent at time $t$. Then, our goal is to compute the Nash equilibrium trajectories for all agents. The trajectories for the human agents are predictions, while the trajectories for the AVs can be executed in a receding horizon planning loop. Finally, none of the agents are assumed to follow constant velocity models.

%% file: img/overview/overview.tex
\begin{figure*}
    \centering
    \includegraphics[width=\textwidth]{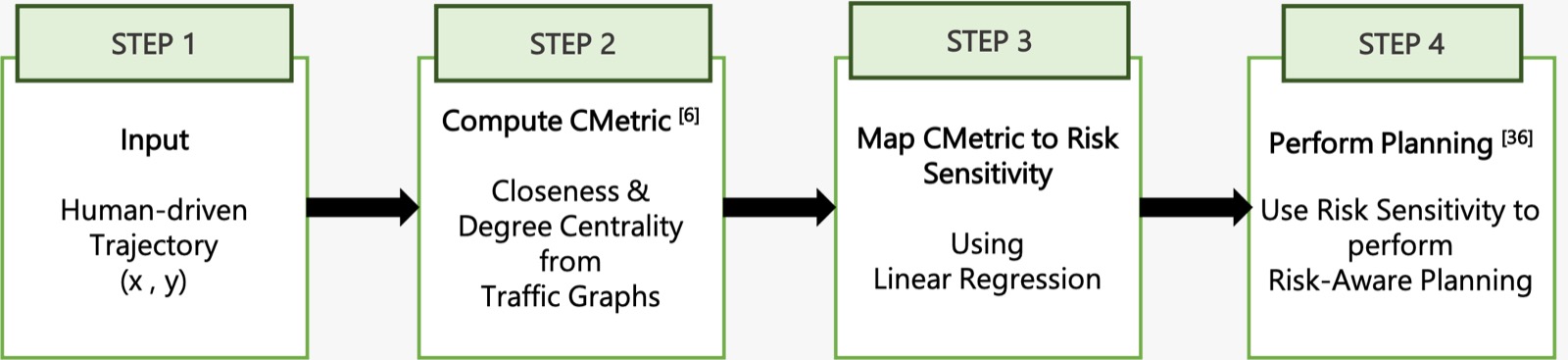}
    \caption{\textbf{Overview:} We outline the steps of our approach for risk-sensitive planning with human agents. \textit{(Block 1)} In the first step, humans drive a vehicle in an OpenAI simulator using a keyboard. \textit{(Block 2)} The next step is to compute the CMetric value for each human user that encodes the aggressive (or conservative) nature of the human driver via certain indicators such as speeding, overtaking, and zigzagging through the simulated traffic. \textit{(Block 3}) We map the human drivers' behaviors to their risk sensitivity using a linear transformation. \textit{(Block 4)} Finally, based on the risk sensitivity, we perform game-theoretic risk-aware planning using the planner developed by Wang et al.~\cite{wang2020game}.}
    \label{fig: overview}
    \vspace{-10pt}
\end{figure*}

%% file: Sections/4-approach.tex
\section{Algorithm}
\label{sec: gameplan}

We describe our algorithm (Figure~\ref{fig: overview}). The first step is to read the trajectories for every agent over a finite horizon $T$, denoted by $\Xi_T$. These trajectories correspond to human agents. The second step is to compute the CMetric value, $\zeta$, for each agent during $T$; the CMetric value encodes the aggressive (or conservative) nature of the driver via certain indicators such as speeding, overtaking, and zigzagging. The third step consists of mapping an agent's CMetric to their risk sensitivity. This is performed using a linear transformation obtained by simple linear regression. This is discussed in detail in Section~\ref{subsec: cmetric_to_theta}. Finally, based on the risk sensitivity, we perform game-theoretic risk-aware planning using the planner developed by Wang et al.~\cite{wang2020game}.

\subsection{CMetric to Risk Sensitivity}
\label{subsec: cmetric_to_theta}

We denote the risk sensitivity parameter by $\theta$. We first compute a linear mapping $\mc{M}:Z \longrightarrow \Theta$. Since both $\zeta \in Z$ and $\theta \in \Theta$ are scalars, we can use simple one dimensional linear regression to estimate $\mc{M}$. We create a training dataset by first generating trajectories corresponding to a fixed array of risk sensitivity values ranging from $-5.0$ (risk-seeking) to $+5.0$ (risk-averse). We denote these risk sensitivity values as $\hat\theta$ to indicate they are training values. We then evaluate the CMetric values, also represented using $\hat \zeta$, corresponding to each of these trajectories using the algorithm described in the previous section. The risk sensitivity and CMetric pair constitute the training dataset on which we apply linear regression to estimate linear coefficients $\beta_0$ and $\beta_1$. $\mc{M}$ is then defined as follows,

\begin{equation}
    \mc{M}(\zeta) = \beta_1\zeta + \beta_0,
    \label{eq: mapping}
\end{equation}

\noindent where $\zeta$ is the CMetric value of a human agent at test time.

\subsection{Risk-Aware Planning}

The system dynamics for each agent are given by,
\begin{equation}
\label{eq:pre_oc_lineardynamics}
    x_{t+1} = A_{t}x_{t} + B^{1}_{t}u_{t}^{1} + B_{t}^{2}u_{t}^{2} + w_{t}.
\end{equation}

\noindent To simplify notation, we describe a two-player system, although our approach can easily generalize to $n$ agents. $x_{t} = [x^{1}_{t}, x^{2}_t] \in X$ represents the system state and $x^{i}_{t} = [p^i_x, p^i_y, v^i_x, v^i_y]$ denotes the position (in meters) and velocity (in meters/second) of an agent. $u^1_{t} = a_1, u^2_{t}=a_2 \in U$ are the control inputs for both agents denoting the acceleration of both agents, $w_{t} \sim \mathcal{N}(0, W_{t})$ is the system noise, and $A_{t}, B^{1}_{t}$, $B^{2}_{t}$ are fixed matrices of appropriate dimensions.
An agent incurs the following cost during a finite horizon $T$:
\begin{equation}
\begin{split}
    \Psi^{i} = 
    & \sum_{t=0}^{T-1}\left[
    \frac{1}{2}x_{t}^{T}Q_{t}^{i}x_{t} + l_{t}^{iT}x_{t} + \frac{1}{2}\sum_{j}u_{t}^{jT}R_{t}^{ij}u_{t}^{j}
     \right] + \\ 
    & \frac{1}{2}x_{T}^{T}Q_{T}^{i}x_{T} + l_{T}^{iT}x_{T},
\end{split}
    \label{eq: cost_function}
\end{equation}

\noindent where $Q_{t}\succeq 0$ and $R_{t} \succ 0$. To model risk, we use the exponential risk cost function used in~\cite{wang2020game},

\begin{equation}
    J(\Psi) = \frac{1}{\mathcal{M}(\zeta)}\log \mathbb{E} \left[ e^{\left( \mathcal{M}\left(\zeta\right) \Psi\right)} \right]= R_{\mathcal{M}(\zeta)}(\Psi),
    \label{eq: risk_cost_function}
\end{equation}

\noindent where $\mathcal{M}(\zeta)$ is the risk tolerance of a human driver.

\noindent \textit{\textbf{Remark $1$:} The difference between Equation~\ref{eq: risk_cost_function} and the risk cost function described in~\cite{wang2020game} is that the risk parameter in the latter work includes a fixed value for every agent, whereas in this work, we automatically generate the risk parameter for human agents in a data-driven fashion.}

The optimal strategies for each player can be obtained by minimizing $J(\Psi^i)$ for each agent $i$ and obtaining the Nash equilibrium using Riccati recursion~\cite[Chap. 6]{basar1998dynamic}. However, Equation~\ref{eq: risk_cost_function} is constrained by the fact that $\mathcal{M}(\zeta)$ is bounded. If the $\mathcal{M}(\zeta)$ is too low or too high, then the cost function value approaches $\infty$, also known as ``neurotic breakdown''~\cite{wang2020game}. Due to the data-driven nature of Equation~\ref{eq: risk_cost_function}, in order to ensure optimality, certain traffic parameters such  as traffic density is assumed, since they affect the CMetric value~\cite{cmetric}, and by Equation~\ref{eq: mapping}, the risk sensitivity of the human agent used in Equation~\ref{eq: risk_cost_function}.

%% file: Sections/6-experiments.tex
\section{Experiments and Results}
\label{sec: experiments_and_results}

In this section, we present the results of extensive experiments testing the accuracy of the linear mapping between human driver aggressiveness and risk tolerance. We then evaluate the emergent behaviors associated with the final trajectories generated by the iterative risk sensitive game theoretic solver, compare with~\cite{wang2020game}, which is chosen as the baseline (in which human driver behavior is ignored), and finally, discuss using alternative human driver behavior models. All experiments are performed using a $12$-core $2.60$GHz Intel i7 processor. We conduct open-loop tests that follow the pipeline outlined in Figure~\ref{fig: overview}. We use the OpenAI traffic simulator~\cite{leurent2019social} to compute the CMetric values representing human driver behavior and the python-based controller provided by Wang et al.~\cite{wang2020game} to generate the final trajectories based on the risk tolerances obtained from the corresponding CMetric values. The configurations of both the simulator and the controller (which include the dynamics of the vehicles, traffic density, number of lanes etc.) are kept identical so that all vehicles generated using the controller are tracked in the simulator. 

We compute the CMetric values of the human driver in a highway scenario since we require a fixed duration of time ($5$s) during which we must observe the vehicle's trajectory and its interaction with other vehicles. For the risk-aware trajectory controller, we consider a merging scenario where a human agent must merge onto a highway with another human agent in the target merging lane. Here, the human agent is equivalent to an agent whose risk sensitivity value is obtained from the CMetric value. We assume vehicles follow the center line in their current driving lanes and only consider the vehicle's speed to finish the merging maneuver. In other words, we assume a steering controller will be executed separately for each car to remain in its lane. 

\subsection{Verifying the accuracy of $\mathcal{M}$}
\label{subsec: results_subsection}

\label{subsubsec: accuracy_of_M}

\begin{figure}[t]
    \centering
    \begin{subfigure}[h]{\columnwidth}
      \includegraphics[width=\columnwidth]{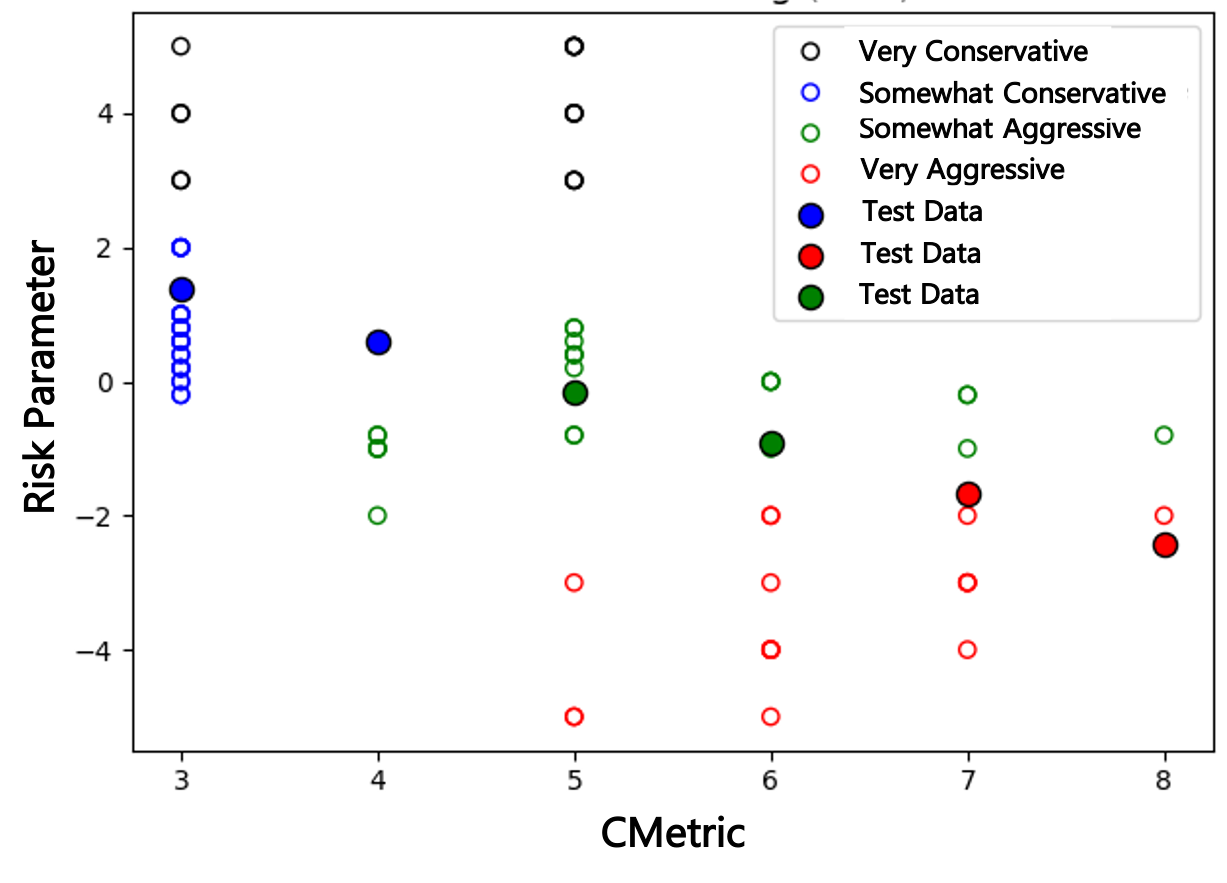}
    \caption{\textbf{KMeans clustering:} As human driver behavior becomes more aggressive (higher CMetric value), the risk parameter $\theta$ tends to decrease. This trend is consistent with the definition of risk sensitivity in~\cite{wang2020game}. Furthermore, our approach can effectively categorize the risk sensitivity of new human drivers (solid colored points) regardless of traffic density, number of lanes, etc.}
    \label{fig: kmeans}
    \end{subfigure}
    \begin{subfigure}[h]{\columnwidth}
  \includegraphics[width=\columnwidth]{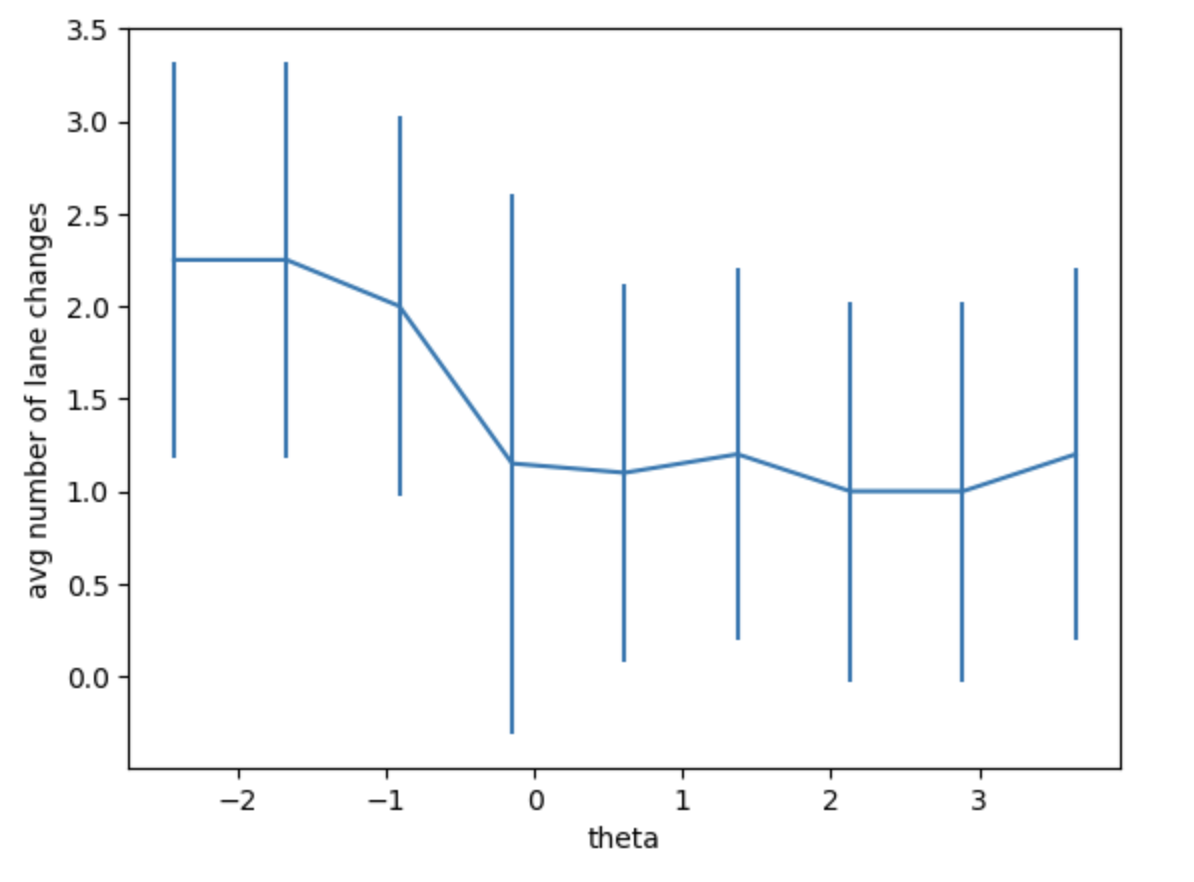}
    \caption{\textbf{Number of lane changes:} Aggressive drivers ($\theta < 0$) yield a greater number of lane changes than conservative drivers ($\theta > 0$).}
    \label{fig: lane_changes}
    \end{subfigure}
    \caption{We highlight the relationship between the CMetric value and the risk parameter $\theta$. Refer to Section~\ref{subsubsec: accuracy_of_M} for further details.}
    \label{fig: kmeans_lane_changes}
    \vspace{-10pt}
\end{figure}

In Figure~\ref{fig: kmeans}, we plot the risk parameter $\theta$ (y-axis) obtained from a given CMetric value (x-axis) via the linear mapping. When computing the risk parameters corresponding to each CMetric value, we vary the simulation configuration (traffic density, number of lanes etc.) to include a range of environments. This results in $\theta$ belonging to a range (as opposed to a fixed value). This is desirable since, in practice, the traffic will vary according to place and time. The risk parameters are clustered into four categories: ``very conservative'', ``conservative'', ``aggressive'', and ``very aggressive''. Each cluster is identified by a color. The empty circles are training data. The goal of this experiment is to cluster a test set of CMetric values (solid-colored points) based on their risk sensitivity. The test data are generated by a human driving the OpenAI simulator~\cite{leurent2019social} in a randomly selected environment consisting of eleven vehicles and four lanes. The results (Figure~\ref{fig: kmeans}) demonstrate that given the CMetric value, the linear regression mapping can accurately identify the risk sensitivity among a wide range of traffic configurations.

\begin{figure}[t]
\centering
\begin{subfigure}{.9\columnwidth}
  \centering
  \includegraphics[width=\columnwidth]{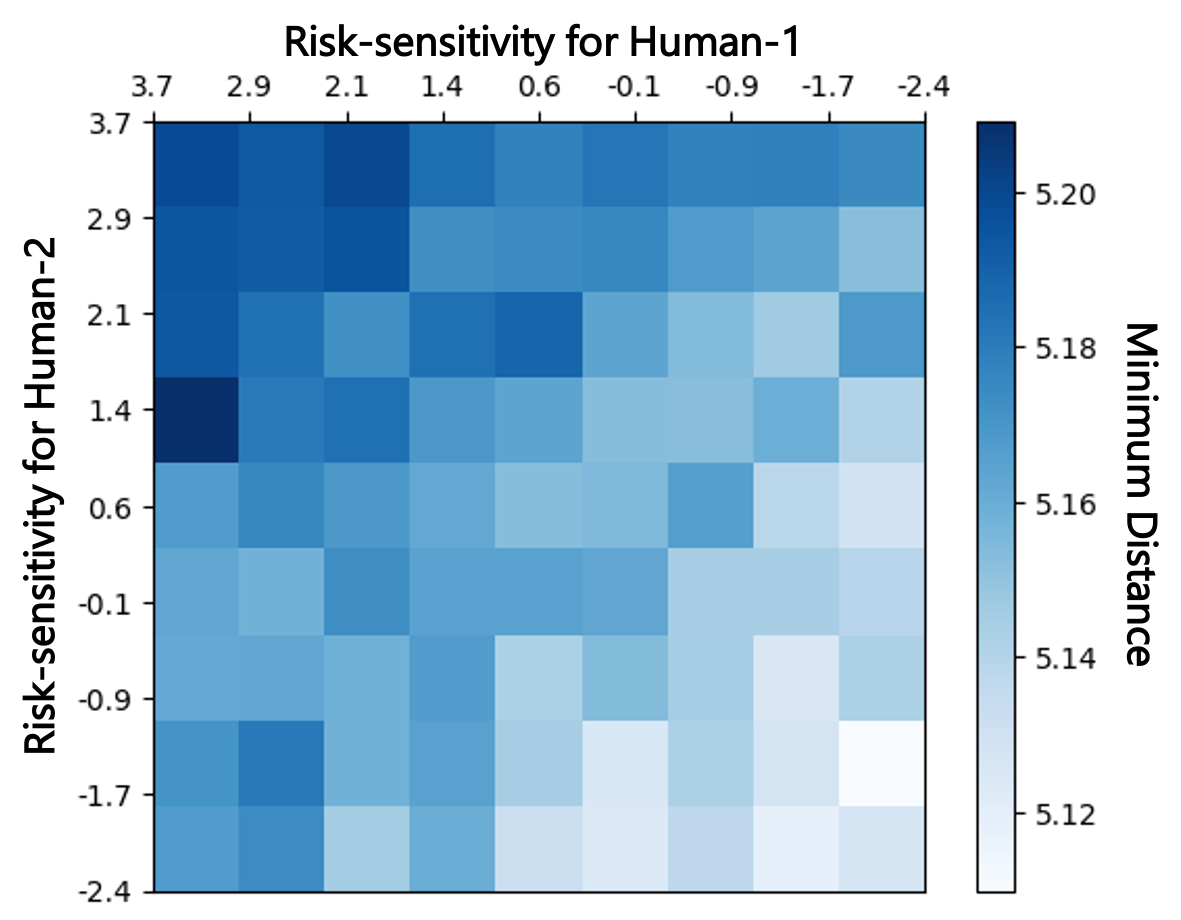}
  \caption{\textbf{Minimum distance:} Darker colors (indicating larger minimum distance) corresponding to two risk-averse human agents and lighter colors (indicating smaller minimum distance) corresponding to two risk-seeking human agents. }
  \label{fig:merge_distance}
\end{subfigure}
\vfill
\begin{subfigure}{.9\columnwidth}
  \centering
  \includegraphics[width=\columnwidth]{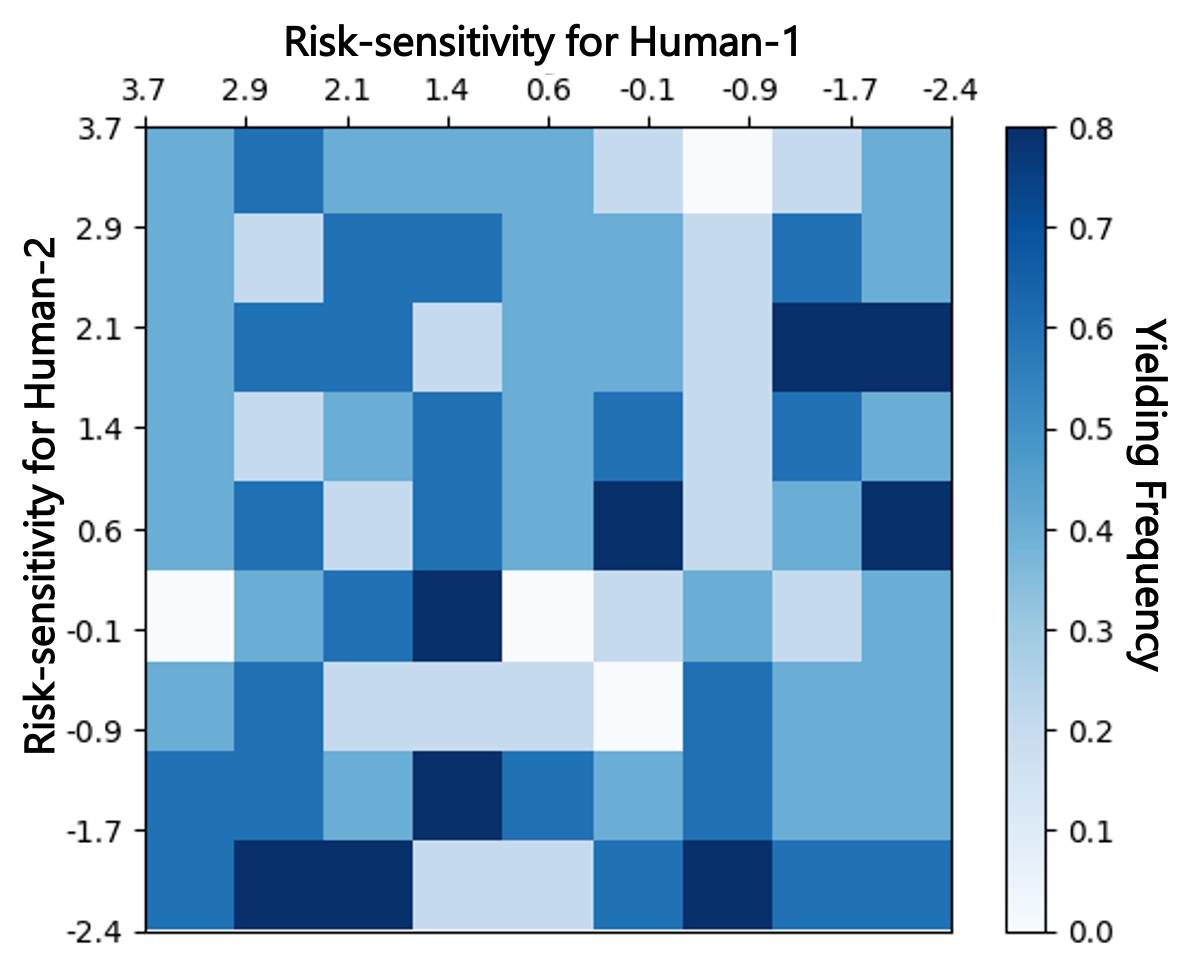}
  \caption{\textbf{Yielding behaviors:} Darker colors (indicating higher likelihood of yielding) corresponding to interactions between a risk-seeking agent and risk-averse agent. As the risk tolerances for both the human drivers are data-driven, and therefore noisy, both agents are adapting to the other. As a result, when either agent is risk-averse, we see higher yielding likelihood (darker colors).  
  }
  \label{fig:merge_yield}
\end{subfigure}
\caption{We demonstrate a range of emergent behaviors. The relative risk sensitivity between the two human agents determines the minimum distance between the two agents and the yielding outcome.}
\vspace{-10pt}
\end{figure}

Another metric we use to validate the correlation between the CMetric and corresponding risk sensitivity is the average number of lane changes. Based on the final trajectories generated from the risk sensitivity parameter (obtained from corresponding CMetric values) and using the controller provided by~\cite{wang2020game}, we measure the average number of lane changes made by the ego-vehicle. The reason for using average number of lane changes as a metric is that aggressive drivers change lanes more frequently than non-aggressive and conservative drivers. The aim of the experiment, therefore, is to check if an aggressive human-driven vehicle (modeled using the keyboard of the OpenAI simulator) results in more lane changes by the final simulated ego-vehicle  (simulated using the python controller) and conversely, if a conservative human driver results in fewer simulated lane changes. In Figure~\ref{fig: lane_changes}, we confirm this is indeed the case; aggressive drivers ($\theta < 0$) yield a greater number of lane changes than conservative drivers ($\theta > 0$).

\input{img/user_study/us1}

\subsection{Emergent behaviors}
\label{subsubsec: quantitative_final_trajectories}

We evaluate the final trajectories generated using the learned risk sensitivity of human drivers in a merging scenario where a human agent attempts to merge onto the highway. Different risk sensitivities yield a range of emergent behaviors. For example, in~\cite{wang2020game}, Wang et al. showed that two risk-averse agents maintain a larger minimum distance between them, while risk-seeking agents may allow a smaller gap. Further, in an interaction between a risk-averse and a risk-seeking agent, there is a higher likelihood of the risk-averse agent yielding to the risk-seeking agent.

The experiments conducted by Wang et al. modeled synthetic agents for which the risk sensitivity must be manually chosen. Here, we run the same set of experiments for human agents. In Figure~\ref{fig:merge_distance}, we can observe darker colors (indicating larger minimum distance) corresponding to two risk-averse human agents and lighter colors (indicating smaller minimum distance) corresponding to two risk-seeking human agents. In Figure~\ref{fig:merge_yield}, we can observe darker colors (indicating a higher likelihood of yielding for the risk-averse agent) corresponding to interactions between a risk-seeking agent and a risk-averse agent. 

\subsection{User studies}
\label{subsubsec: qualitative_final_trajectories}

We recruited $27$ participants to respond to a user study consisting of two questions. The first question involved showing two video clips of final trajectories. The first clip (top) consisted of a risk seeking trajectory ($\theta= -2.429$), while the second (bottom) consisted of a risk averse trajectory ($\theta = 3.651$). Participants were not told the risk preferences that generated the trajectories, and were asked to identify which trajectory corresponded to an aggressive driver. The goal of this question is to qualitatively assess the emergent nature of the final trajectories. That is, based on simply observing the nature of the trajectory, can a human distinguish between the generated trajectories? We answer in the affirmative; $26$ out of the $27$ participants correctly identified the risk seeking driver as the aggressive driver.


\subsection{Comparing with the baseline}
\label{subsubsec: Comparing_with_baseline}
    \begin{figure}[t]
      \centering
      \includegraphics[width=\columnwidth]{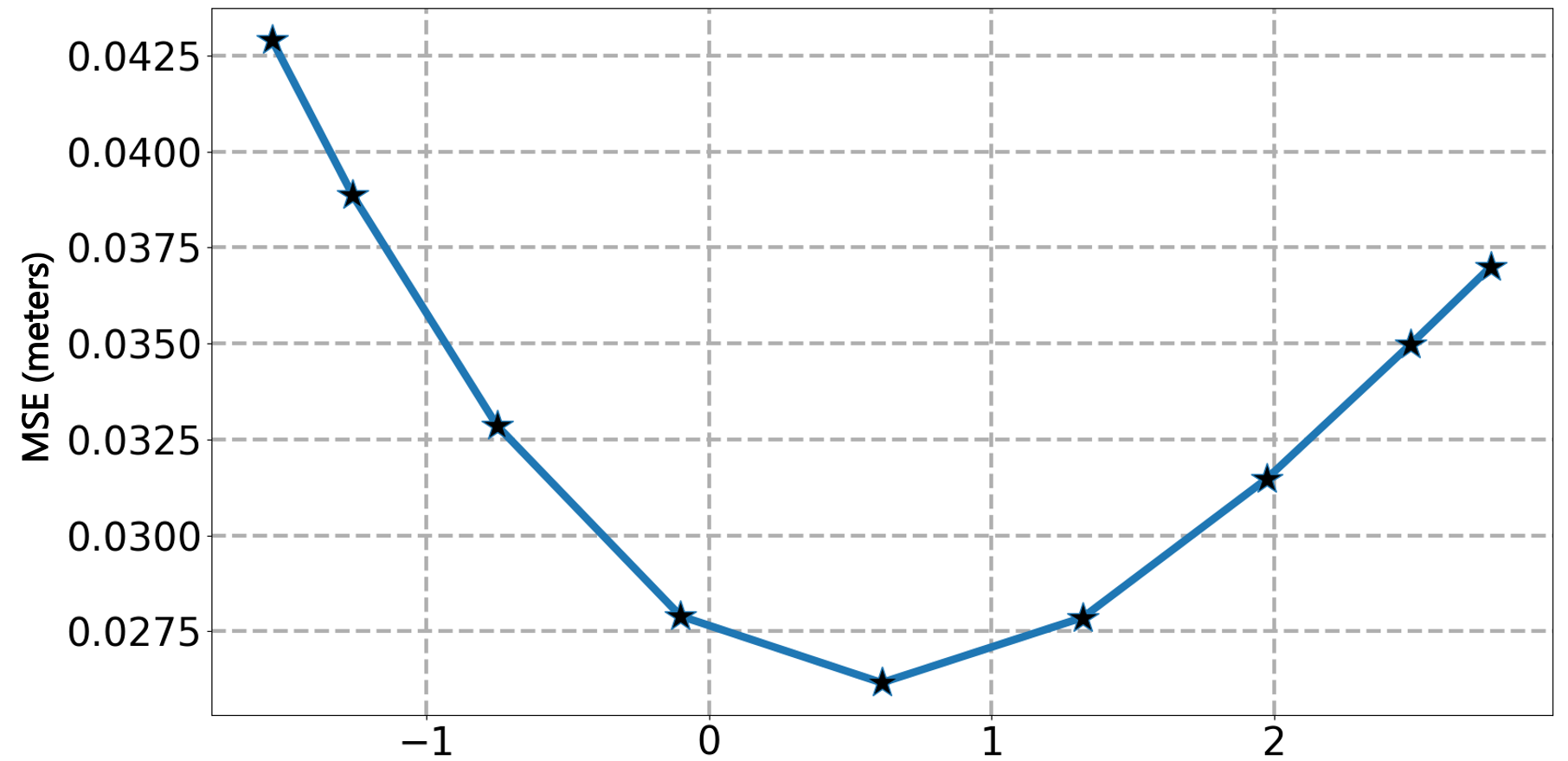}
    \caption{\textbf{Comparison with~\cite{wang2020game}:} The approach by Wang et al. assumes a neutral risk sensitivity for human agents. However, when an ego-agent interacts with a human driver who may be aggressive or conservative (indicated by the negative and positive values on the x axis, respectively), then the assumption of neutral risk tolerance results in an error in terms of absolute value of minimum relative distance between the two agents. See Section~\ref{subsubsec: Comparing_with_baseline} for more details.
    }
    \label{fig: error}
\vspace{-10pt}
\end{figure}

We compare our modified risk-aware planner with existing planners that do not model human drivers and show that modeling human drivers results in safer navigation. Specifically,~\cite{wang2020game} (and similar planners) assign a fixed neutral risk tolerance to human drivers and the ego-vehicle generates to the human driver accordingly. There are two outcomes:

\begin{enumerate}
    \item Suppose the human driver is, in fact, aggressive. Then, by modeling the driver with a neutral risk tolerance, the ego-vehicle may stray close to the aggressive driver as opposed to keeping a safe distance from them.
    
    \item Conversely, suppose the human driver is conservative. Then, by modeling the driver with a neutral risk tolerance, the ego-vehicle may enter a brief deadlock during which both agents wait to see who moves first.
\end{enumerate}

We aim to capture these inefficiencies via a single error metric, which is the absolute value of the minimum relative distance between the two agents. This metric is ideal since in both cases, it measures the discrepancy between the expected distance and the actual observed distance. For example, we show that in the first case, the expected minimum relative distance between both agents is more than the observed distance while in the second case, we show the observed minimum distance is more than the actual distance. In both cases, the error is positive by virtue of the absolute value. Empirically, the maximum RMSE observed is $0.0425$m or $10\%$ as shown in Figure~\ref{fig: error}.


\subsection{Using alternative human driver behavior models}
\label{subsubsec: using_alternate_models}

Thus far, we have successfully demonstrated that CMetric can be effectively integrated with risk-aware planning to generate game-theoretic behavior-rich trajectories. Alternative models for human driver behavior such as the SVO can theoretically be used. However, there are practical issues when it comes to integrating these alternative models in risk-aware planning. Here, we discuss some of these challenges. SVO is an offline technique that requires a large volume of training data to learn a data-driven reward function via inverse reinforcement learning. Our technique is meant to be deployed in realtime and, as such, we test in an open-loop simulation and use \textit{active} metrics such as yield \%, frequency of lane changes, and minimum distance between agents. The SVO approach, on the other hand, uses RMSE to measure the deviation of the prediction trajectories from the ground-truth trajectories. We do not assume the availability of ground-truth data. In future, we will conduct experiments comparing CMetric with SVO once the source code for SVO is public.

%% file: img/user_study/us1.tex
\begin{figure}[!t]
\centering
\begin{subfigure}{\columnwidth}
  \centering
  \includegraphics[width=\columnwidth]{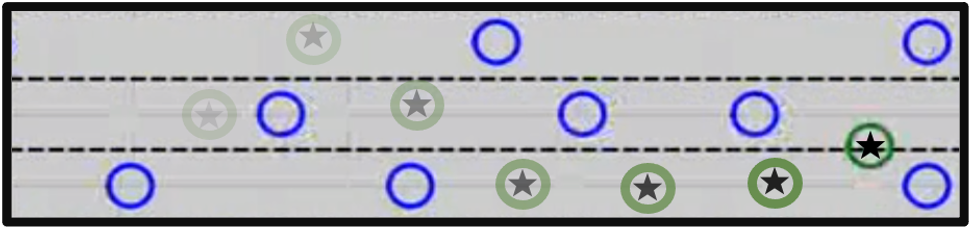}
  \caption{\textit{(top)} Risk-aware planner generates an aggressive trajectory for the green agent consisting of zigzagging, overtaking, and overspeeding. Black stars indicate the positions.}
  \label{fig:merge_diagram1}
\end{subfigure}
\vfill
\begin{subfigure}{\columnwidth}
  \centering
  \includegraphics[width=\columnwidth]{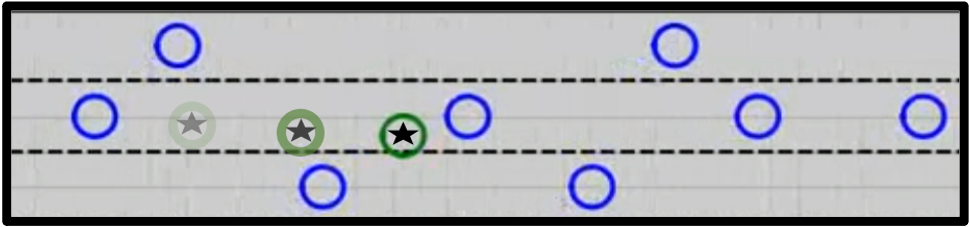}
  \caption{\textit{(bottom)} Risk-aware planner generates a conservative trajectory where the green agent prefers to stay in the center lane instead of switching to the top lane, which has sufficient space for overtaking the blue agent in front. Black stars indicate the positions.}
  \label{fig:merge_diagram2}
\end{subfigure}
\begin{subfigure}{\columnwidth}
  \centering
  \includegraphics[width=\columnwidth]{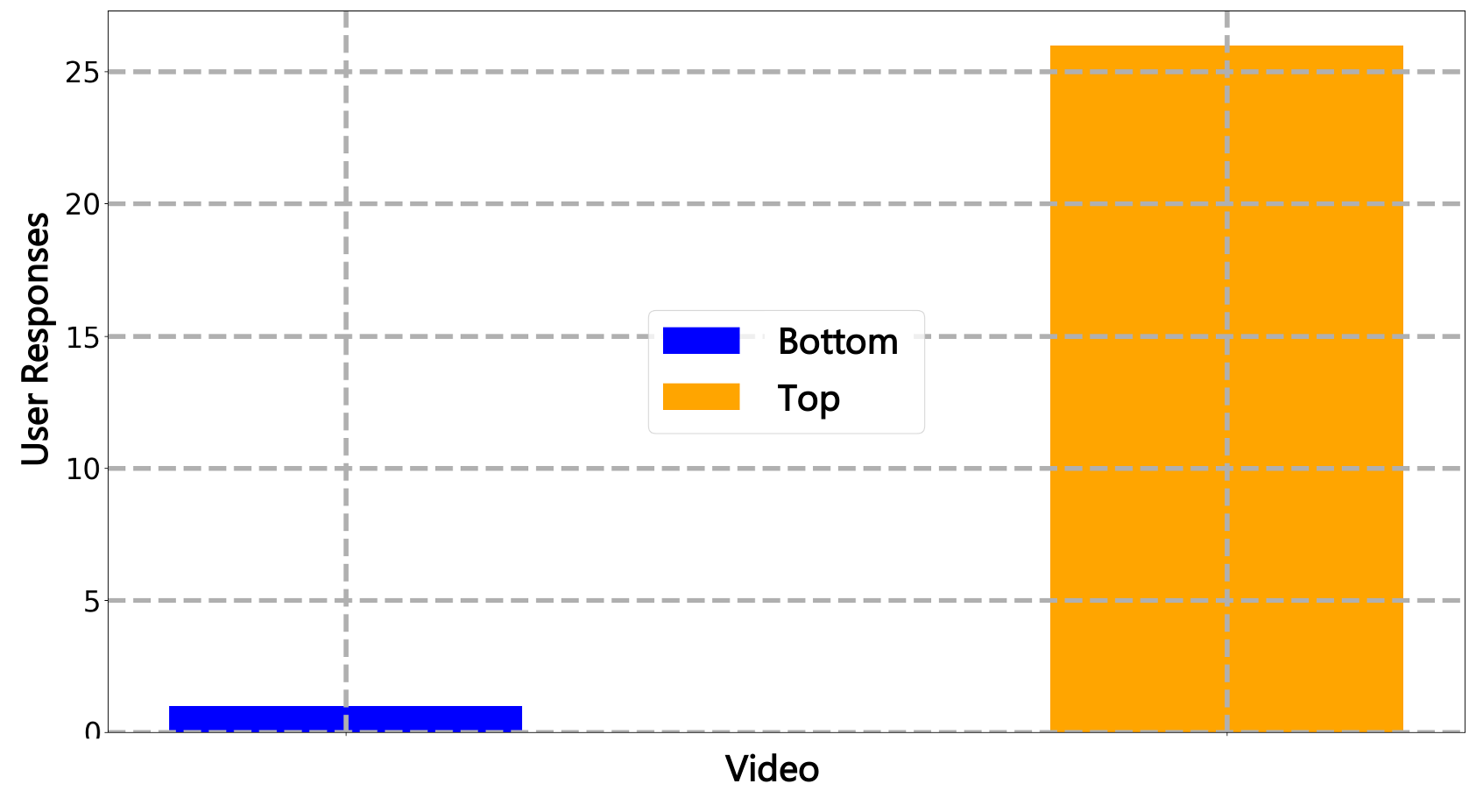}
  \caption{\textbf{Q. Indicate which agent (top vs. bottom) is more aggressive.}}
  \label{fig:merge_diagram2}
\end{subfigure}
\caption{\textbf{Qualitative analysis}. We conduct a user study to qualitatively assess the emergent nature of the final trajectories. $26$ out of the $27$ participants correctly identified the aggressive driver.}
\label{fig:merge_diagram}
\vspace{-10pt}
\end{figure}

%% file: Sections/7-conclusion.tex
\section{Conclusion, Limitations, and Future Work}
\label{sec: conclusion}

We presented an approach for risk-aware planning in multi-agent traffic with human agents. The basic intuition of our approach is that aggressiveness of a driver is linearly correlated with risk preference. That is, aggressive drivers are more risk-seeking while conservative drivers are more risk-averse. Accordingly, we integrate a human driver behavior model~\cite{cmetric} with the risk-aware dynamic game solver in~\cite{wang2020game} via simple linear regression to derive a mapping between driver behavior and risk tolerance. Our results show that aggressive human driving results in more frequent lane-changing. 
We show that conservative drivers generally yield to aggressive drivers while maintaining a greater distance from them. Finally, we confirm that the final trajectories obtained from the risk-aware planner generate emergent behaviors though a comprehensive user study in which participants were able to distinguish between aggressive and conservative drivers.

There are some limitations to our method. Currently, we have tested our approach in an open-loop simulation where we use two different simulators for the human behavior model and the trajectory planner. To use both simulators in open-loop simulation effectively, the environment configuration must be kept identical, which is cumbersome and a hindrance. In the future, we will explore a closed-loop simulator that combines the human behavior model and the risk-aware trajectory planner.